# Risk Assessment of Autonomous Driving: Integrating Technical Failures, Ethical Dilemmas, and Policy Frameworks

**Boyi Chen*, Shengqin Chu*, Zicheng Wang*, Brian Baetz and Zhen Gao**

Faculty of Engineering, McMaster University
E-mails: chenb66@mcmaster.ca; chus25@mcmaster.ca; wang1476@mcmaster.ca; baetz@mcmaster.ca; gaozhen@mcmaster.ca

*These authors contributed equally to this work

**Abstract:**

Autonomous driving technology has the potential to reduce the large number of road traffic accidents caused by human error each year, but it also brings new types of risks that need to be evaluated from the aspects of technology, ethics and regulations. Based on public crash data from the National Highway Traffic Safety Administration (NHTSA), disengagement reports from the California Department of Motor Vehicles (DMV), the MIT Moral Machines dataset, and a comparative regulatory analysis of five jurisdictions, we have found that the main types of technical failure modes are perception and classification errors. These account for a relatively large proportion of the reported accidents, and it can be concluded that there are different ethical frameworks for autonomous vehicle decision-making, and inconsistent regulations in different areas increase the uncertainty of widespread application. Generally speaking, the problems of technology, ethics and regulation are closely related and need to be solved together. Therefore, this paper recommends a more adaptive and cooperative governance approach that combines engineering standards, ethical discussion, and institutional supervision.

## 1. Introduction

Autonomous driving technology development is one of the most important tech evolutions in our modern transportation lifetime. Over the past 10 years, companies including Waymo, Cruise and Tesla, along with legacy automotive manufacturers have spent tens of billions attempting to build what are essentially vehicles that can handle complex road conditions on their own. The global market for autonomous vehicles has been estimated to be in the range of $54 billion as of 2025, with forecasts suggesting a valuation exceeding $300 billion by 2030 (Statista, 2024) Nevertheless, there are outstanding risks in three key areas: technical, ethical, as well as regulatory that need to be seriously evaluated until we can be assured of the widespread introduction or use of autonomous vehicles.

As a result, the motivation for such assessment is both urgent and comprehensive. To be specific, the promise of autonomous driving is anchored in safety: human error is responsible for an estimated 94 percent of serious motor vehicle crashes in the United

States (NHTSA, 2023). Worldwide, the World Health Organization estimates approximately 1.35 million human deaths each year due to road traffic injuries with tens of millions more suffering non-fatal injuries that often result in long-term disability (World Health Organization, 2023). Supporters of self-driving vehicles point to how algorithmic systems are less fallible than human drivers, and believe that they could reduce these numbers dramatically; saving hundreds of thousands of lives every year. This utilitarian calculus, fewer deaths, fewer injuries, fewer economic losses, is the moral and economic basis for this technology.

However, the deployment of AVs does not just remove risk, instead, it reallocates and transforms how that necessary risk is expressed. New failure modes from self-driving systems, such as sensor degradation, software errors, adversarial vulnerabilities, algorithmic bias and unpredictable interactions with human road users pose classes of risk that are qualitatively different from the familiar ones from human drivers. If technology is to redeem its safety promise and not become a novel, potentially systemic source of harm, then risks must be quantified, and the emergent risks mitigated by regulation or practice as soon as possible.

The most pressing are issues surrounding technical failure. Autonomous vehicles rely on a set of sensors, consisting of cameras, LiDAR (light detection and ranging), radar and ultrasound. Each one has its own limitations. A tragedy of cascading perception failures was enacted in the October 2018 fatal crash of an Uber ATG test vehicle in Tempe, Arizona (National Transportation Safety Board 2019). In addition to perception, the sensitivity of deep neural networks to adversarial perturbations adds a cybersecurity aspect that has yet been confronted by current architectures (Goodfellow et al., 2015).

Autonomous driving leads to questions of nature both utterly unique and moral. An algorithm of how the system would assign blame is much like a computational version of an ancient moral philosophy problem: A trolley is headed toward five people on the tracks, but it can be diverted onto another track where there's only one person. While the MIT Moral Machine experiment, which surveyed millions of participants from 233 countries, showed that moral preferences in regard to such scenarios vary widely across cultures and suggest a single universal ethical framework is not likely to be compressed into AV decision-making systems (Awad et al., 2018). Historical Note: One reason for studying ethics is that ethical issues extend beyond traffic accident scenarios to encompass algorithmic bias, including significant differences in pedestrian detection based on gender or race, as noted in certain studies (Wilson et al., 2019), as well as issues such as surveillance, data privacy, the displacement of professional drivers by autonomous vehicle technology, and unequal access to autonomous mobility services.

There are many separate regulations. Although the United States has a complex combination of federal regulations and state laws, it is trying to build a harmonized type-approval system, such as that found in Europe. China is taking an even more top-down approach and has introduced committed pilot zones. There is a lack of unified international norms, and thus, problems in liability have arisen among producers.

Given the current situation of intertwined technical, ethical and regulatory problems, this paper aims to achieve three goals. First, by way of an organized study, the main technical fault reasons in autonomous driving systems will be identified, and based on this, deficiencies in the current perception and decision pathways will be identified through actual crash data and disengagement reports. Second, it examines the moral problems that arise during the operation of autonomous vehicles and explores ethical systems that have been proposed to address these issues in actual practice, as well as their cultural backgrounds. Third, it compares the governance systems of the United States, countries in the European Union, China and Japan, etc., and points out some small differences that are quite significant, as well as possible ways to build more consistent governance systems.

An all-encompassing method is necessary because the problems that need to be solved in terms of technology, ethics and laws will not be resolved individually. Operating a technically safe vehicle under faulty regulations is still dangerous for society, and although there is a belief among some people who support strict regulations that technology will solve all problems in time, there are still certain fundamental technical limitations that innovation alone cannot address. This paper aims to offer a more all-round view of autonomous driving risk by integrating the results from all these areas, rather than the individual siloed views that will need to be combined in order to fully understand this technology.

The rest of the paper is organized as follows: Section two reviews the related literature in the four different fields; the framework and data sources are outlined in Section three; Results of technical performance, ethical trade-offs and regulatory comparison are presented in Section four; Cross-domain interactions are discussed in Section five; Limitations and future research opportunities are mentioned in the concluding remarks.

## 2. Literature Review

With the rapid development of autonomous driving technology, many areas in engineering, computer science, philosophy, law and public policy have been affected and numerous studies have been conducted. The four areas that provide the intellectual basis for the present risk assessment are introduced in this section in the previous studies.

### 2.1 Foundations of Autonomous Vehicle Safety Research

The first study of autonomous vehicle safety has laid the foundation for our understanding of the problem and established the requirements for a certain level of system reliability. One of the first technical introductions to the development of robotic cars, which described the architecture of sensors, probabilistic reasoning methods and machine learning techniques at that time, also appeared in a short tutorial provided by Thrun (2010) based on the experience of competitors in the DARPA Grand Challenge and Urban Grand Challenge. His work indicated that the autonomous driving problem is in fact a problem of uncertainty-robust perception and planning, which remains one of the focuses of technical research to this day.

Kalra and Paddock (2016) were one of the first to point out this problem in the RAND Corporation's report: statistically speaking, it would be impossible to demonstrate that autonomous vehicles are safe on public roads. It is expected that a fleet of autonomous vehicles will need to drive tens or even hundreds of millions to hundreds of billions of miles before it can be shown with statistical certainty compared with human drivers. Therefore, in order to meet the demands of simulation-based testing and formal verification in the structure of safety cases, new validation means have appeared. The Kalra and Paddock study is one of the frequently cited works in the field of AV safety to this day, and it continues to provide a reference for the minimum acceptable evidentiary standard for deployment.

Koopman and Wagner (2017) have taken the above as a foundation to address the problem of safety verification for large-scale advanced systems based on machine learning. They thought that the current automotive safety standards, such as ISO 26262, were based on the deterministic behaviour of traditional cars and thus not suitable for systems where decisions are made by deep neural networks with billions of parameters. The framework they proposed includes ODD specification, edge-case identification methodology and continuous post-deployment monitoring; these ideas have been employed by the leading organizations and regulatory authorities to this day.

Also in 2017, Shalev-Shwartz, Shammah and Shashua put forward an official mathematical safety model called Responsibility-Sensitive Safety (RSS), and it has been demonstrated that under the above assumptions, the planning decisions of an autonomous vehicle will never lead to a dangerous situation.

RSS is a way to give humans a reasonable assurance of safety, otherwise it would only be a statistical statement from a machine learning system. Although it is well-known, critics have pointed out that the assumptions made about the behaviour of other road users in this model may be too simple and cannot fully consider all circumstances of mixed-traffic conditions (Pek, Manzinger, Koschi, & Althoff, 2020).

### 2.2 Technical Failure Modes and Perception Vulnerabilities

A considerable amount of research has been carried out on the particular technical failure modes of AV. Goodfellow, Shlens and Szegedy (2015) were the first to introduce the idea of adversarial examples, which are small additions to the input data that can cause neural networks to make mistakes with high confidence. Research has shown that even a small change or a small disturbance in an image can be recognised by a good AI classifier. Therefore, we are not sure if the vision-based perception system for autonomous driving will be robust in terms of safety.

Eykholt and others (2018) have shown that the above problems actually exist in the real world; that is, a small addition to the road sign is prone to making the deep learning model identify a stop sign as a speed limit sign during actual operation. It can be seen from this paper that the perception pipeline of autonomous vehicles is vulnerable to damage by malicious attacks and other natural visual noise.

Research on the problems of sensors has only recently begun. Petit, Stottelaar, Feiri and Kargl (2015) have found that low-cost laser equipment can be used to spoof LiDAR systems and add false objects to the environmental model of a vehicle. Cao and others (2019) have developed some high-tech LiDAR spoofing attacks that can deceive multi-sensor fusion systems, and it has been shown that even with the use of multiple types of sensors, adversarial threats cannot be fully prevented.

Amodei and others (2016) have investigated the problem of distributional shift, which is the difference between the deployment environment and the training data; they have also put forward the five required directions for AI safety: prevention of adverse consequences, avoidance of reward hacking, scalable oversight, safe exploration and robustness to distributional shift. The structure of this model is applicable to self-driving cars, but since the training data are mostly in good weather conditions, they may perform poorly in adverse weather and low-light situations.

In the 2019 investigation by the NTSB into the 2018 Uber ATG accident, some detailed information about the cause of this technical failure chain was released to the public. According to the report, although the perception system detected the presence of a pedestrian about six seconds before a collision, it kept changing the classification of the object among "vehicle", "bicycle" and "unknown", and thus neither the prediction module nor the planning module was able to generate a reasonable avoidance plan. According to the report, Uber has disabled the Volvo's factory-installed emergency braking system and also has a problem with the safety culture.

**2.3 Ethical Frameworks for Algorithmic Decision-Making**

There are many ethical problems in self-driving cars that have drawn much attention from academia and society. Bonnefon, Shariff and Rahwan (2016) were among the first to conduct empirical research on public attitudes towards self-driving cars in the early 2010s. In light of the social dilemma in their results, although most people were willing to accept the idea of utilitarian self-driving cars that would sacrifice the convenience of passengers to save more pedestrians, users were found to be highly reluctant to ride in self-protecting vehicles. Therefore, there is a fundamental problem for both the producer and the government.

The MIT Moral Machine experiment (Awad et al., 2018) took this issue to a different level and collected roughly 40 million answers from people around the world; it has since been published in Nature. The results show that there is a general moral tendency to sacrifice human life for the lives of animals or the many for the sake of the few, and there are also significant cross-cultural differences and prejudices against certain groups based on age and social status. Based on this data, it has been concluded that there is no single ethical system for AV that can be universally accepted; thus, it is difficult to form a uniform standard for the world.

Nyholm and Smids (2016) have pointed out in their studies that the framing of the trolley problem is not suitable for the circumstances of actual driving; for example, there is a lack of certainty in the results, it is a continuous problem over time rather than a sudden two-choice situation, and it generally falls under probabilistic risk management rather than deterministic division of harm. It has been proposed that to achieve the ethical design of self-driving cars, the corresponding risks should be distributed among all road users equitably, and there should be no specific rules dictating a binary sacrifice scenario.

Lin (2016) put forward the idea that AV ethics should not only address serious traffic accidents. Many technical design decisions, such as how to urgently brake, how far back to keep from a following vehicle, and how to navigate around cyclists, are in fact moral choices about the degree of risk that different road users are willing to take. The design decisions in daily life that affect the distribution of risk for all people are more serious than occasional dilemma situations.

In 2017, the German Ethics Commission on Automated and Connected Driving released its first set of governmental ethical guidelines; it listed 20 basic principles, such as that the accident algorithm should not be based on people's personal attributes, human life must be given priority over property damage, and the party that approves risk-generating technology should bear more responsibility than the person only using it. These are the reasons why Europe has built its regulations.

**2.4 Policy and Governance Frameworks**

Fagnant and Kockelman (2015) were the first to study AV widely and found that full-scale implementation of self-driving cars in the United States would save around $1.3 trillion each year in economic losses from accidents, production and energy costs, etc. However, they have also pointed out some problems with the policy, such as liability division, cybersecurity, data privacy, etc.

Glancy (2015) believes that the federal government should set up vehicle safety standards, but provinces and municipalities are also responsible for dealing with registration, licensing, traffic laws, etc.; otherwise, if each place has its own rules, the situation will become even more chaotic. It should be pointed out that the EU AI Act (2024) considers autonomous vehicles to be high-risk AI systems and will be subject to strict conformity assessment, human supervision, etc. (European Parliament 2024). Stilgoe (2018) believes that, in a responsible manner, the development of autonomous driving technology should address both the risks for people's lives and the values of society in the process.

Marchant and Lindor (2012) had predicted that there would be problems with tort liability and expected that the existing product liability system, covering design defects and failure to warn, would need to be reformed for autonomous vehicles. Geistfeld (2017) proposed a system based on the negligence standard, that is to say, instead of the traditional reasonable-person standard for human drivers, the reasonable operation of a properly functioning autonomous system should be used.

In short, this literature review shows that the development of technology has exceeded the progress of ethical agreement, regulations and empirical safety data, which is the reason for conducting this comprehensive analysis now.

## 3. Analytical Framework

The two ways for analysing and comparing the risks of autonomous driving in the three areas in this paper will be secondary data analysis and structured comparative review. Unlike the first type of empirical study, we will use public data and regulations from all over the world to learn general rules, exceptions and other changes.

### 3.1 Research Approach

At the same time, we will conduct an analysis of the technical performance data, the ethical framework and the regulatory comparison in a convergent manner and then integrate the results through an overall discussion. A typical way of safety engineering to display the flow of harm after a hazard event is the Bow-tie risk model (de Ruijter & Guldenmund, 2016); thus, we will use it to organise the connections among technical failures (on the left), serious accidents in safety (in the middle), and ethical and regulatory responses (on the right).

### 3.2 Illustrative Hypothetical Example: Siloed versus Holistic Risk Assessment

To illustrate the value of an integrated risk assessment approach, here is a simplified autonomous driving scenario. An autonomous vehicle is operating at night in an urban environment and approaches a pedestrian crosswalk. The perception system detects an object near the vehicle's path, but low illumination and partial occlusion reduce classification confidence. As a result, the system is unable to determine with sufficient certainty whether the object is a pedestrian, a stationary object, or an unknown obstacle. The delayed interpretation of the scene leads to a late response and a near-collision event.

In this situation, a siloed assessment would primarily focus on the immediate technical failure. The incident may be attributed to limitations in low-light perception, insufficient training data, inadequate sensor fusion, or inappropriate classification thresholds. While these technical improvements are important, they address only one dimension of the problem. An ethical assessment would emphasize the need to protect vulnerable road users, particularly pedestrians. However, ethical objectives cannot be effectively implemented if the technical system is unable to reliably identify vulnerable individuals in the first place. Similarly, a regulatory assessment might require incident reporting or impose restrictions on night-time operation, yet such measures may be insufficient without a clear understanding of the underlying technical causes.

In contrast, a holistic assessment considers the interactions among technical, ethical, and regulatory factors. From a technical perspective, the incident highlights the need for improved low-light perception, uncertainty-aware decision-making, and fail-safe braking strategies when classification confidence is low. From an ethical perspective, the system should adopt precautionary behaviour when vulnerable road users may be present, even under conditions of uncertainty. From a regulatory perspective, the event suggests the need for clearly defined operational design domain limits, reporting requirements for safety-critical near-miss events, validation procedures for perception-related software updates, and transparent liability allocation among developers, operators, and vehicle owners.

This example demonstrates why autonomous driving risks should not be evaluated solely through technical, ethical, or regulatory lenses in isolation. The same event can simultaneously represent a perception failure, a failure to adequately protect vulnerable road users, and a governance failure if testing, reporting, or accountability mechanisms are inadequate. By integrating these perspectives, a holistic assessment can identify not only the immediate causes of a risk event, but also the broader safeguards required to reduce the likelihood and severity of similar incidents in future deployments.

**3.3 Data Sources**

Based on the following public data, we have conducted the analysis:

- NHTSA Standing General Order (SGO) data (2021–2024): Mandatory crash reports for vehicles with ADS and ADAS operating in the United States.
- California DMV Autonomous Vehicle Disengagement Reports (2020–2023): Annual filings submitted by all companies testing AVs on California public roads.
- MIT Moral Machine dataset (Awad et al., 2018): More than 40 million ethical preference decisions by respondents in 233 countries.
- Regulatory texts: Primary legislation and regulation from the United States, European Union, China, Japan, and the United Kingdom.

**3.4 Analytical Methods**

To learn about the technical risks, we will collect and organize the accident data in NHTSA SGO reports and CA DMV disengagement filings to find out why the vehicles failed to work correctly and whether there is an increasing problem. We then calculate the compound annual improvement rate of the data for long-term disengagement prediction. According to the NHTSA Traffic Safety Facts and FHWA Traffic Volume Trends, we have set the performance of autonomous vehicles as the reference for human-driver crash baselines.

For the ethical evaluation, all the above frameworks will be employed to compare the concepts of utilitarianism, deontology and Rawlsian justice in five dimensions: self-consistency, public acceptance, legality in accordance with the law, ease of calculation and cultural relevance. The foundation of the above indicators is the literature on computational ethics, and they have been supported by empirical results from the published Moral Machine dataset (Awad et al., 2018).

To help in the analysis of the policies, we have constructed a six-dimensional comparison matrix of the regulatory standards for safety certification, liability distribution, data governance rules, ethical code implementation, establishment of an enforcement mechanism and adaptability to change; then, deficiencies, conflicts and exemplary cases will be extracted from the five sections that cover different regulatory systems.

The three strands of analysis have been combined to form a Bow-tie diagram, and it is shown that there are main deficiencies in technical risk, ethics and regulations, thus elevating the overall level of risk.

## 4. Findings and Analysis

### 4.1 Technical Failure Modes

An analysis of NHTSA's Standing General Order collision reports covering the period from July 2021 to November 2023 noted that vehicles equipped with Advanced Driver Assistance Systems operating on public roads in the United States alone were involved in a total of more than 2,400 reported accidents, while vehicles equipped with Fully Automated Driving Systems were involved in nearly 340 accidents (NHTSA, 2024).The bulk of Level 2 reports were attributed to Tesla's ADAS, which is more indicative of fleet size than necessarily higher per-mile risk. As a matter of fact, Waymo and Cruise are operating in larger deployments, it made sense that they reported a higher raw number of incidents than other ADS operators.

According to the base level of human-driver baseline crashes in NHTSA data, there are approximately 6.7 million police-reported crashes per year for about 3.2 trillion vehicle miles travelled (VMT), which is about 2.1 crashes per million VMT. The accident rate is about 1.34 per 100 million miles (2022 data) (NHTSA, 2023; FHWA, 2024).

Table 1: U.S. Motor Vehicle Crash Statistics - Human Baseline (2018–2023) (NHTSA, 2023; FHWA, 2024; National Safety Council, 2024)

| Year | Total Crashes (millions) | Fatalities | VMT (trillions) | Fatal Rate per 100M VMT | Injury Rate per 100M VMT |
|---|---|---|---|---|---|
| 2018 | 6.73 | 36,560 | 3.24 | 1.13 | 78.0 |
| 2019 | 6.76 | 36,096 | 3.23 | 1.11 | 77.0 |
| 2020 | 5.25 | 38,824 | 2.83 | 1.37 | 72.0 |
| 2021 | 6.10 | 42,939 | 3.15 | 1.36 | 76.0 |
| 2022 | 6.20 | 42,795 | 3.26 | 1.31 | 75.5 |
| 2023 | 6.15 | 40,990 | 3.30 | 1.24 | 74.0 |

*Sources: NHTSA Traffic Safety Facts; FHWA Traffic Volume Trends; National Safety Council Injury Facts (2024 edition). 2023 figures are preliminary estimates.*

In 2020-2021, a problem arose: Although VMT decreased by about 12.7% due to the impact of the COVID-19 pandemic, the number of fatalities increased by 6.2%, and thus the fatal rate rose from 1.11 to 1.37, an increase of 23.4%; this can be traced back to some people driving in dangerous ways (such as high speed, not wearing a seatbelt and impaired driving) in a small number of cars. It is a problem of context; autonomous driving will not act this way in all sorts of traffic situations, and the structural safety advantages of autonomous driving will be underestimated by basic statistics.

Data also show that although there have been continuous, albeit slow, improvements in the safety of vehicle engineering (airbags, electronic stability control, automatic emergency braking) for decades, the human fatal crash rate has been around 1.1-1.4 per 100 million miles since 2014 and has not shown any improvement. Therefore, there is little room for improvement in the reduction of road traffic accidents by current means; fundamental modifications to human-driven operation will be necessary to promote the development of autonomous vehicles.

**4.2 ADS Reliability Trajectories**

The California DMV disengagement data is one of the most reliable publicly available long-term sources for understanding the reliability of autonomous driving systems.

Table 2: ADS Disengagement Rates - California DMV Reports (2020–2023) (California Department of Motor Vehicles, 2021; 2022; 2023; 2024)

| Company | Year | Autonomous Miles | Disengagements | Miles/Disengagement | YoY Improvement |
|---|---|---|---|---|---|
| Waymo | 2020 | 628,839 | 21 | 29,945 | — |
| Waymo | 2021 | 2,325,190 | 63 | 36,908 | +23.2% |
| Waymo | 2022 | 4,091,783 | 84 | 48,712 | +32.0% |
| Waymo | 2023 | 7,139,492 | 119 | 59,996 | +23.2% |
| Cruise | 2020 | 770,049 | 27 | 28,520 | — |
| Cruise | 2021 | 876,104 | 23 | 38,092 | +33.6% |
| Cruise | 2022 | 1,532,531 | 36 | 42,570 | +11.7% |
| Cruise | 2023 | 2,076,240 | 54 | 38,449 | -9.7% |
| Pony.ai | 2020 | 199,507 | 21 | 9,500 | — |
| Pony.ai | 2021 | 305,617 | 24 | 12,734 | +34.0% |
| Pony.ai | 2022 | 510,923 | 17 | 30,054 | +136.0% |
| Pony.ai | 2023 | 612,100 | 13 | 47,085 | +56.6% |

*Sources: California DMV Autonomous Vehicle Disengagement Reports (2020–2023). Cruise 2023 data may reflect operational changes following October 2023 suspension.*

Waymo has made some improvements and increased the number of miles per disengagement from about 30,000 to around 60,000 over the past four years, rising by about 19% each year. However, as Kalra and Paddock (2016) have shown, it is not necessary to travel about 11 billion miles before being 95% confident that AVs are only 20% safer than human drivers. Waymo has gone over 40 million miles, and that is not a large number. The factors of improvement are (1.45x - 2.79x) and they are logarithmic, not exponential; that is to say, the marginal difficulty increases with the age of the system, and it is a "long-tail problem".

One of the most thorough safety evaluations of autonomous vehicles is Waymo's peer-reviewed safety report (Kusano et al., 2023, Nature Medicine), which examined 7.14 million miles of passenger-only driving data. The rate of police-reported autonomous vehicle accidents was 85% lower than the human-driven benchmark, and the rate of accidents resulting in injuries was 57% lower. Additionally, there were no at-fault accidents that resulted in injuries to cyclists or pedestrians throughout the entire dataset. Although these findings point to encouraging opportunities for more automation, their statistical strength is still constrained. In terms to prove the statistical superiority of autonomous driving in terms of fatalities, hundreds of millions more miles of data are still required, since the 95% confidence upper bound for zero fatalities over 7.14 million miles is roughly 0.42 fatalities per 100 million vehicle miles travelled (VMT).

Based on internal simulation replay data disclosed by Waymo in its Voluntary Safety Self-Assessment Report (Waymo, 2023), industry estimates suggest that without human intervention, approximately 2% to 8% of critical disengagement events would result in a collision. From another perspective, this probability indicates that Waymo’s current rate of one disengagement per 60,000 miles is equivalent to a potential accident occurring once every 750,000 to 3 million miles driven. However, this figure remains lower than the accident rate of 2.1 per million vehicle miles traveled (VMT) for human drivers, although its confidence interval remains wide.

#### 4.3 Ethical Framework Trade-offs

There are basic trade-offs between the ethical frameworks for autonomous driving decision-making, and no single framework clearly outperforms the others in terms of all evaluation criteria, according to our methodical comparison.

**Deontological (rule-based) approaches** offer high internal consistency and computational tractability. For instance, absolute prohibitions such as “never target a specific individual” can be easily translated into deterministic decision-making logic. That said, in certain situations, strict adherence to rules can lead to counterintuitive outcomes, as minor rule violations might actually prevent significant overall harm. This has resulted in lower public acceptance of this approach in research studies (Lin, 2016).

**Utilitarian approaches** have greater popular support. According to the Moral Machine dataset, aggregate damage minimization mirrors prevailing lay moral intuitions worldwide (Awad et al., 2018). However, there exists significant cultural variety in utilitarian calculus because the number of people, age, and social status are all weighted differently in different places, which makes it difficult for manufacturers to use the same algorithms globally.

**Fairness-based Rawlsian approaches** that prioritize the protection of the most vulnerable road users are aligned with equity concerns, but are computationally intractable, as real-time vulnerability assessment requires contextual inference beyond the capabilities of current perception.

Operational data indicate the “trolley problem” cases where the driver has to choose between two unfavourable outcomes are less common than regular operational failures, such as sensor occlusion, mapping errors, and prediction errors. Based on engineering analyses, slowdown, course modification, or controlled stopping can resolve more than 99.5% of safety-critical scenarios without necessitating forced harm-allocation judgments (Nyholm & Smids, 2016). This outcome has practical ramifications, because it implies that the trolley-problem scenarios receive a disproportionate amount of public and scholarly attention compared to their operational significance, and that engineering resources are best used to eliminate the routine perception and prediction failures that account for the great majority of real-world AV safety events.

Ordinary engineering decisions, such as speed setting, following distance arrangement, interaction protocols with pedestrians, and the degree of lane changing, have a far more pressing impact on the risk characteristics of autonomous driving systems than those two rare challenges. These “micro-ethical” decisions, as Lin (2016) expressed, are often invisible to public scrutiny but have a disproportionate impact on safety outcomes. A vehicle designed to maintain tighter following distances in the interest of maximizing traffic throughput implicitly accepts a higher risk of rear-end collisions. A vehicle designed to be very conservative in yielding to pedestrians at any potential crossing point may be safer, but at the expense of traffic throughput and user frustration that threatens adoption.

### 4.4 Regulatory Framework Comparison

The five areas show different results in the comparison.

Table 3: Comparative Regulatory Frameworks for Autonomous Vehicles (2024) (NHTSA, 2023; European Parliament, 2024; UNECE, 2021; UK Parliament, 2024; Deloitte, 2024)

| Dimension | United States | European Union | China | Japan | United Kingdom |
|---|---|---|---|---|---|
| Primary Legislation | No federal AV law; state patchwork | EU AI Act + UNECE R157 | National Innovation Plan; pilot zone rules | Road Traffic Act Amendment (2023) | Automated Vehicles Act 2024 |
| Max Permitted Level | Level 4 (state-dependent) | Level 3 (≤60 km/h, ALKS) | Level 4 (designated zones) | Level 4 (designated routes) | Level 4 (authorized entities) |
| Liability Model | Product liability + state tort | Manufacturer at Level 3+ | Operator/platform in zones | Owner (L3); Operator (L4) | Authorized Self-Driving Entity |

| Dimension | United States | European Union | China | Japan | United Kingdom |
|---|---|---|---|---|---|
| Ethical Framework | No binding requirements | Ethics-by-design (AI Act) | State guidance | SAE-J ethical guidelines | No value-of-life hierarchy |
| Deployment Scale | ~1,500 robotaxis (Waymo) | Limited pilots | ~3,000+ robotaxis (Baidu, Pony.ai) | Honda L3 small volume | Pilots only |

*Sources: NHTSA; EU AI Act (2024/1689); UNECE R157; UK Automated Vehicles Act 2024; Deloitte Global Automotive Regulatory Guide (2024).*

This comparison allows a number of analytic observations. First, the liability question is the pivot of divergence in policy.

Some analysis has been conducted on the above. Firstly, the problem of liability is the cause of the differences in policies. The UK model has the best architecture; it established a new type of legal entity, the "Authorised Self-driving Entity", to take on liability and resolve the ambiguity of who is responsible among manufacturers, operators and users in other countries. The US model is still the most dispersed, and the liability outcome depends on the circumstances of the accident, the level of technology for the product, and existing product liability rules.

Secondly, the system of China is state-led. There is no corresponding norm in the Western system for the obligation to promptly submit mobile data to the government's monitoring platform; in general, they also have different assumptions about the link between the mobility data of individuals and state supervision. Although it is the case that there is a large central safety data set, and this type is relatively simple for the regulatory authorities to learn from, it also has problems in supervision and data sovereignty.

Thirdly, the speed restriction of the EU (up to 60km/h for Level 3 ALKS) is the most conservative technical limit among the main jurisdictions, and this is due to the precautionary principle in European regulatory culture. Therefore, Level 3 is only for traffic congestion on the expressway; although it is safe, it will not offer more experience to improve the reliability of the system.

Fourthly, in places with compulsory accident reporting, such as the US (NHTSA SGO) and China, a large number of safety reports have also been collected from other parties that have volunteered to do so. Without such data, it is not possible to establish a rule of law based on evidence; thus, this requirement of transparency in the regulations should be met as advocated by Stilgoe (2018).

**5. Discussion: Cross-Domain Risk Interactions**

The main conclusion of the analysis is that the three types of risks, technical, ethical and policy, are not isolated but rather work together. There are other reasons that a siloed assessment method may be less than desirable.

**Technical Limitations in Ethical Realisation**. The limitation of technical perception is an inhibition of ethics' application. A utilitarian harm-minimisation algorithm will not be able to distinguish between a child and a traffic cone at a distance of 50 meters in low light because of errors in the perception system. According to the analysis of disengagement data, both perception and classification errors are still the main types of failures in the leading systems; therefore, the necessary technical ability for advanced ethical reasoning in edge cases is still many years away. Following Nyholm and Smids' (2016) idea, engineering effort is to correct habitual perception errors and only change behaviour in very rare cases of dilemma.

More rules are riskier. Because there are no uniform international rules, some companies have chosen to set up their production bases in places with lax regulation and lower costs; as a result, there will be an increasing trend of a "race to the bottom" in terms of regulations and ethics. The EU is relatively strict about the maximum speed of 60km/h for Level 3, and the US has different rules at the state level. Companies may choose to manufacture in an area with a low barrier, gain operating experience there, and then adapt this experience to develop various goods around the world; thus, the looser regulation in that production jurisdiction has implicitly set an international standard.

**Correction of Investment as a Market Signal.** Since 2021, there has been an investment correction and the market has identified its problems; in terms of funding, it has dropped from a high of US$28.4 billion to about US$10-13 billion in 2024 and has declined by more than 57% over time (PitchBook, 2024). Although the problem of technical long-tail, regulatory ambiguity and unanswered questions about the liability of self-driving solutions are still difficult to resolve, the industry has been led by well-capitalised companies with a track record of deployment (Waymo) or by strategic support from the government (Baidu Apollo). Some well-funded competitors have exited the industry (Argo AI closed in 2022, Cruise stopped operating in 2023), and at the same time, there have been technical and regulatory problems.

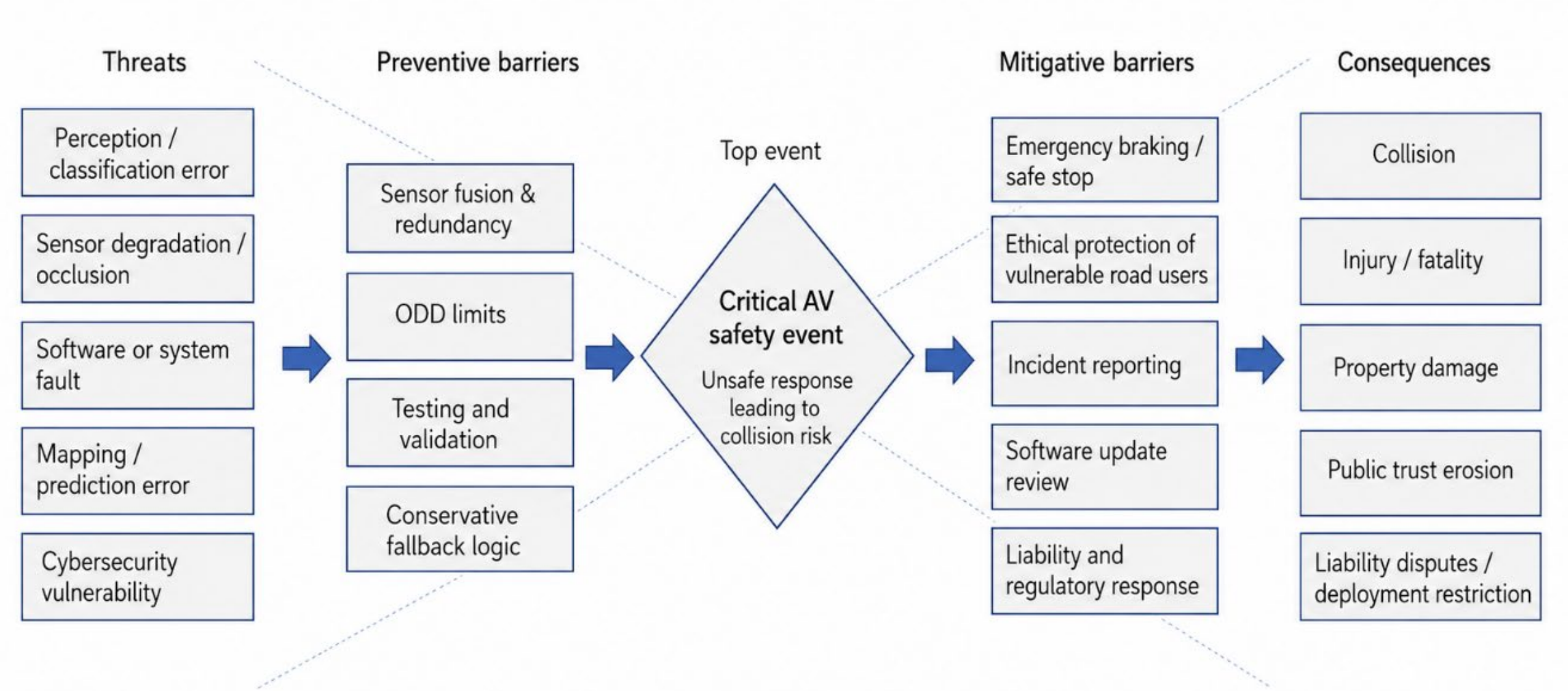

Figure 1: Bow-Tie model for integrated autonomous driving risk assessment. Adapted from de Ruijter & Guldenmund (2016)

**Implications for Governance.** Therefore, in the process of maintaining good governance, all links of the risk system should be covered. As shown in Figure 1, the Bow-Tie model used in this study highlights how technical threats on the left side may develop into a critical AV safety event, and how ethical safeguards and regulatory controls on the right side can mitigate the resulting consequences. This structure show why AV risk should not be evaluated only through a technical lens: ethical safeguards and regulatory controls are also necessary parts of the overall safety architecture (de Ruijter & Guldenmund, 2016); otherwise, both will be absent. For example, if all three areas are weak at the same time, for instance, if the jurisdiction has lowered the requirements for technical validation but failed to establish ethical rules, then the remaining risk to public safety will be much higher. However, the UK has been coordinating the plans of all three in recent years to develop a 2024 integrated Automated Vehicles Act, and although cross-domain governance is feasible, a single plan may be more practical for all parties.

**6. Concluding Remarks**

An analytical review of the risk associated with autonomous driving has been offered in this paper in three interrelated domains: policy frameworks, ethical quandaries, and technical failures. Several important findings emerge from this analysis.

From a technical perspective, perception and classification errors remain the main failure modes in deployed autonomous vehicle systems. Leading platforms have shown steady improvement, with reliability increasing by about 19–25% per year. Waymo's published data also suggests that its property-damage and injury-related crash rates are much lower than human driving benchmarks. However, autonomous vehicles still have far fewer real-world miles compared with the billions of miles needed to make strong statistical claims about fatal-crash safety. Therefore, it is still too early to conclusively say that AVs are safer than human drivers in fatal-crash scenarios (Kalra & Paddock, 2016). The "long-tail" problem also remains a major technical challenge, because rare and unusual driving situations make further reliability improvements increasingly difficult.

The ethical analysis shows that autonomous vehicle decision-making cannot be guided by one universally accepted moral framework. The Moral Machine experiment revealed clear cultural differences in how people judge ethical driving decisions, which makes it difficult to design one ethical algorithm that works equally well across the world. However, since real-life moral dilemma scenarios are extremely rare compared with everyday technical failures, developers should focus more on reducing common perception and prediction errors rather than overemphasizing rare sacrifice-based scenarios.

Analysis of the policy shows that there are many different regulations in various places; they have different standards for responsibility, tests, ethics, etc., making deployment uncertain and susceptible to regulatory evasion. The UK's Approved Self-Driving Entity model is a good architectural idea to solve the problem of liability.

**Limitations.** Several limitations warrant acknowledgment. This study uses secondary data analysis instead of primary experimental research; the authors did not do original sensor testing, simulation modeling or human-subjects experiments first. Second, the publicly available datasets analyzed for this project such as NHTSA SGO, CA DMV reports, have well-known limitations including inconsistent reporting definitions across manufacturers and potential underreporting of minor incidents, and the absence of standardized severity coding. Third, the ethical analysis is primarily situated within Western philosophical frameworks and English scholarship at risk of marginalizing non-Western ethnic traditions that can be highly pertinent in global deployment contexts associated with China or Japan. Fourth, the quick advance of technology implies a few evaluations may lapse as other designs are created. Lastly, this analysis was necessarily broad due to the cross-domain nature of this paper and has resulted in a lack of depth within any individual domain.

Future research should prioritize several critical directions. This makes longitudinal empirical studies that follow safety performance of commercially deployed AVs across regional geographies and operational domains necessary to establish a level of situational awareness sufficiently beyond risk modeling. As they are for aviation and pharmaceutical regulation, the emergence of internationally recognized safety benchmarks to assist with validation (and regulatory consistency) would facilitate progress. To achieve this, ethical frameworks based on genuine stakeholder values variety must be developed through multidisciplinary research that integrates moral psychology, cultural anthropology, and participatory design. AV risk: V2X communication topologies increase the attack surface massively, therefore the cybersecurity aspects of AV risk need significantly more attention. Last but not least, it is important to research adaptive models of regulation to control a technology whose risks and capabilities are always evolving, not static.

In short, autonomous driving is a powerful technology that will change the way people travel in their cars; to realize this ideal responsible, integrated and comprehensive risk governance needs to be established, and the technical problems should not be treated in isolation but rather addressed together as part of a sociotechnical system.

**Acknowledgments**

The authors acknowledge the publicly available data from the National Highway Traffic Safety Administration (NHTSA), the California Department of Motor Vehicles and the MIT Moral Machine project. The authors did not conduct primary experiments and all quantitative analysis in the paper is based on secondary data, as cited throughout the paper. We sincerely appreciate the organizations and institutions that have released the open datasets for this research and analysis.